\documentclass[conference]{IEEEtran}
\IEEEoverridecommandlockouts
\usepackage[utf8]{inputenc}
\usepackage[T1]{fontenc}
\usepackage{graphicx}
\usepackage{subfigure}
\usepackage{stfloats}

\usepackage{cite}
\usepackage{booktabs}

\def\BibTeX{{\rm B\kern-.05em{\sc i\kern-.025em b}\kern-.08em
    T\kern-.1667em\lower.7ex\hbox{E}\kern-.125emX}}

\begin{document}

\title{Manipulation Mask Generator: High-Quality Image Manipulation Mask Generation Method Based on Modified Total Variation Noise Reduction\\
}

\author{\IEEEauthorblockN{1\textsuperscript{st} Xinyu Yang}
\IEEEauthorblockA{\textit{School of Computer Science} \\
\textit{Sichuan University}\\
Chengdu, China \\
2021141460237@stu.scu.edu.cn}
\and
\IEEEauthorblockN{2\textsuperscript{nd} Jizhe Zhou}
\IEEEauthorblockA{\textit{School of Computer Scienc} \\
\textit{Sichuan University}\\
Chengdu, China \\
jzzhou@scu.edu.cn}
}

\maketitle

\begin{abstract}
In artificial intelligence, any model that wants to achieve a good result is inseparable from a large number of high-quality data. It is especially true in the field of tamper detection. This paper proposes a modified total variation noise reduction method to acquire high-quality tampered images. We automatically crawl original and tampered images from the Baidu PS Bar. Baidu PS Bar is a website where net friends post countless tampered images. Subtracting the original image with the tampered image can highlight the tampered area. However, there is also substantial noise on the final print, so these images can't be directly used in the deep learning model. Our modified total variation noise reduction method is aimed at solving this problem. Because a lot of text is slender, it is easy to lose text information after the opening and closing operation. We use MSER (Maximally Stable Extremal Regions) and NMS (Non-maximum Suppression) technology to extract text information. And then use the modified total variation noise reduction technology to process the subtracted image. Finally, we can obtain an image with little noise by adding the image and text information. And the idea also largely retains the text information. Datasets generated in this way can be used in deep learning models, and they will help the model achieve better results.

\end{abstract}




\section{INTRODUCTION}
With the advances in image editing techniques and user-friendly editing software, low-cost tampered or manipulated image generation processes have become widely available\cite{background1}. Non-professional users can easily edit and tamper images without leaving obvious visual traces\cite{background2}. Some tampered images are amusing and harmless, But some can damage the reputation of others and even spread rumors to cause panic. Meanwhile, traditional detection methods play a limited role in processing these images. Therefore, deep learning is more needed in this realm. On the other hand, deep learning models have demonstrated their power in many applications. For example, convolution neural networks (CNNs) achieve promising performance in many computer vision and natural language processing  applications\cite{background3}. 
\\
\indent Deep learning provides a novel approach to identifying features for tempered regions, which inherently represent characteristics of the tempered regions appearing in the dataset\cite{background5}. Compared to traditional image detection methods, deep learning models can automatically learn the features and patterns of images from a large amount of data and can more accurately recognize and classify these features and designs, thus achieving higher accuracy. And they can learn more complex features and patterns, thus better coping with various deformations and distortions in image tampered, improving the robustness and stability of the algorithm. Also, Deep learning models can increase the number of layers and parameters to improve the complexity and performance of the algorithm, thus better adapting to different image-tampered scenarios and application requirements. Besides, Deep learning models automatically warn parameters according to different tampered scenarios and datasets, thus better adapting to various image-tampered detection tasks. Meanwhile, faster processing speed is also one of its advantages: they can use high-performance hardware such as GPUs to accelerate computing, thus achieving faster image-tampered detection speed to meet the requirements of real-time and high efficiency.\\	
\indent	In image-tampered detection, deep learning requires a large number of high-quality datasets for training and validation. Large-scale, high-quality datasets play an essential role in the deep learning era, which act as the catalyst stimulating and accelerating technique development\cite{datasets}. However, Some databases for CMFD(copy-move forgery detection) already exist, but they are not suited for evaluating post-processing methods\cite{background6}. Firstly, current datasets suffer from homogeneity and insufficient sample size, making it difficult for deep learning algorithms to predict new tampered techniques in practical applications. Secondly, existing datasets are often generated through simulated tampered behavior, which may only partially reflect the tampered situation in real-world scenarios. Therefore, collecting and annotating real-world datasets is vital but requires significant time and cost.

\begin{table}[!htbp]
\normalsize
    \caption{The existing tamper detection data sets}\label{tab:001} \centering
    \begin{tabular}{ccccc}
        \toprule[1.5pt]
        Existing DataSets               & Shortcomings\\
        \midrule[1pt]
        Defacto                      &random synthesize\\
        CASIA v1.0                 & monotonous tamper types\\
        NIST 2016                  & label leakage\\
        COVERAGE                   & small sample size\\
        CASIA v2.0                 &unspecified enhancement\\
        \bottomrule[1.5pt]
    \end{tabular}
\end{table}

We list the five main datasets and their shortcomings in Table 1. 
DFACTO is a novel dataset for image and face manipulation detection and localization\cite{dataset0}. 
CASIA v1.0 is one of the more commonly used datasets for image tampered detection, containing a small number of manually tampered images and synthetically tampered images generated by copying and pasting. The database is made publicly available for researchers to compare and evaluate their proposed tampering detection techniques\cite{data1}. However, the sample size is relatively small, and covering all possible tampered scenarios is difficult. NIST 2016 was used in the image tampered detection competition held by the National Institute of Standards and Technology (NIST) in 2016. It contains a large number of tampered images and real-world images, but it is private, and only competition participants can use it, making it difficult for researchers to access. Although COVERAGE includes both manually tampered images and real-world images, the dataset is still being expanded, and the sample size is relatively small, making it challenging to meet the needs of deep learning algorithms. CASIA v2.0 is a large-scale face dataset released by the Chinese Academy of Sciences. The datasets still had some problems, such as imbalanced and limited subject distribution, image quality and metadata, and  Potential for overfitting. These issues may limit the generalization ability of machine learning models trained solely on CASIA v2.0.
We will show our results in Fig.1.
\begin{figure}[h]
\centering
\includegraphics[width=8cm]{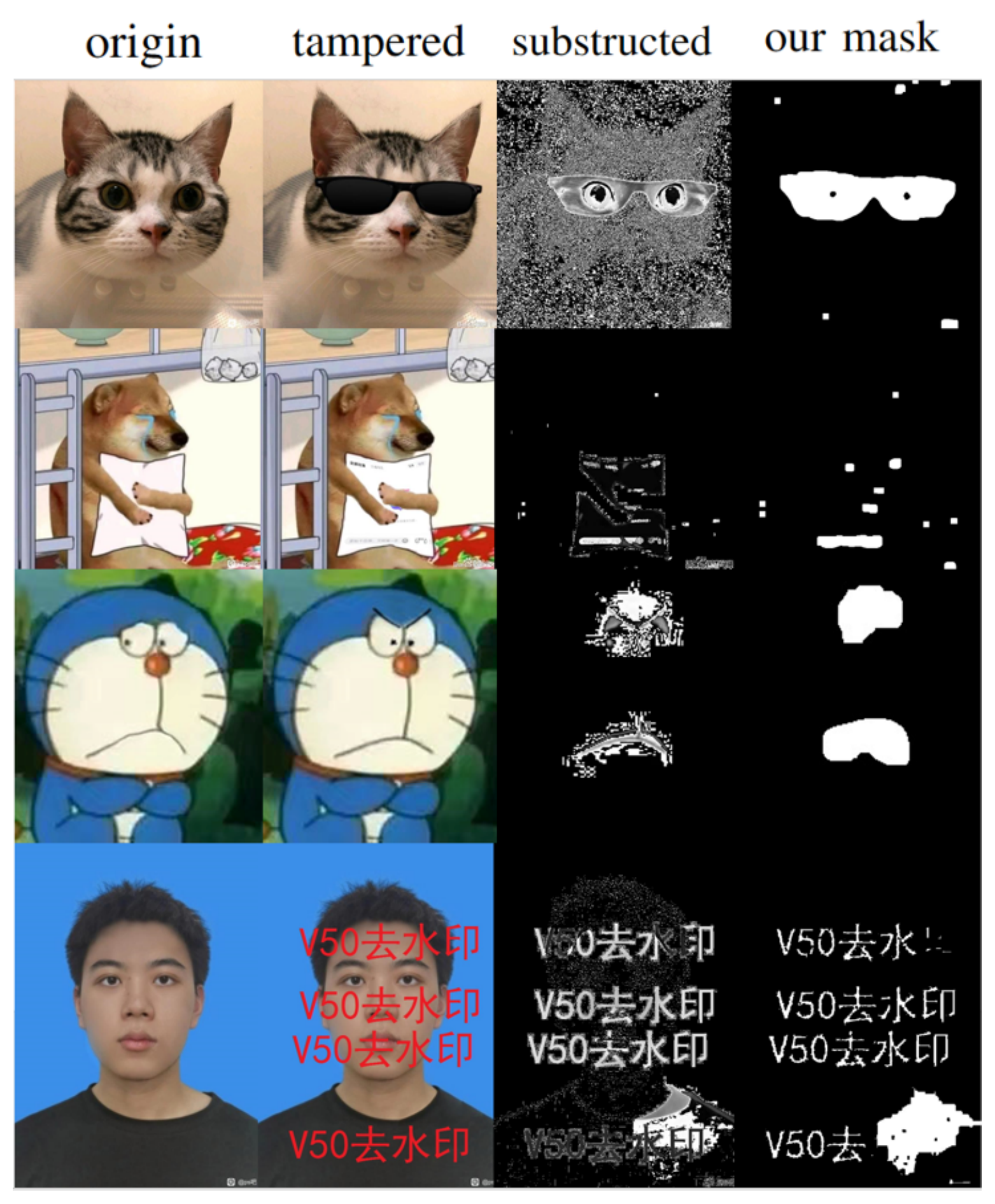}
\caption{We list four groups of images, from left to right, the original image, the tampered image, the noise image after subtracting the two images, and finally, our processing result image. We can see that our method accurately finds the tampered area. The meaning of Chinese characters: pay 50¥ to remove the watermark.
}
\end{figure}

\indent	Considering the increasing demand for public databases for image forensic \cite{data1}, we crawl images from websites with tampered images on the network, such as the Baidu PS Bar. It is a perfect source of original and tampered images. Most users will ask others to help them modify the picture they offer. In that case, Under their posts, there are often a large number of tampered images. We need to save the original and tampered images from different posts. That makes us collect lots of data in a short time. The usual way to find the difference between the tampered image and the original image is to subtract the tampered graph from the original one. However, the user, in the process of modifying the image, tried to make a universe change to the image. That will make the result of subtraction have a lot of noise, even full-screen noise. Using total variation noise reduction can effectively solve this problem. This method combines character recognition technology and some switching operations, which can significantly retain text information and reduce noise. \\	
\indent	In general, the contribution of our paper contains:	
\begin{itemize}
    \item We propose a modified total variation noise reduction method. Total variation noise reduction can remove various types of noise, which will help us obtain a high-quality mask.
    \item We design a way to automatically obtain a large number of datasets, which will help us quickly get a large number of available data.
\end{itemize}

\begin{figure*}[t]
\centering
\includegraphics[width=16cm,height=7.5cm]{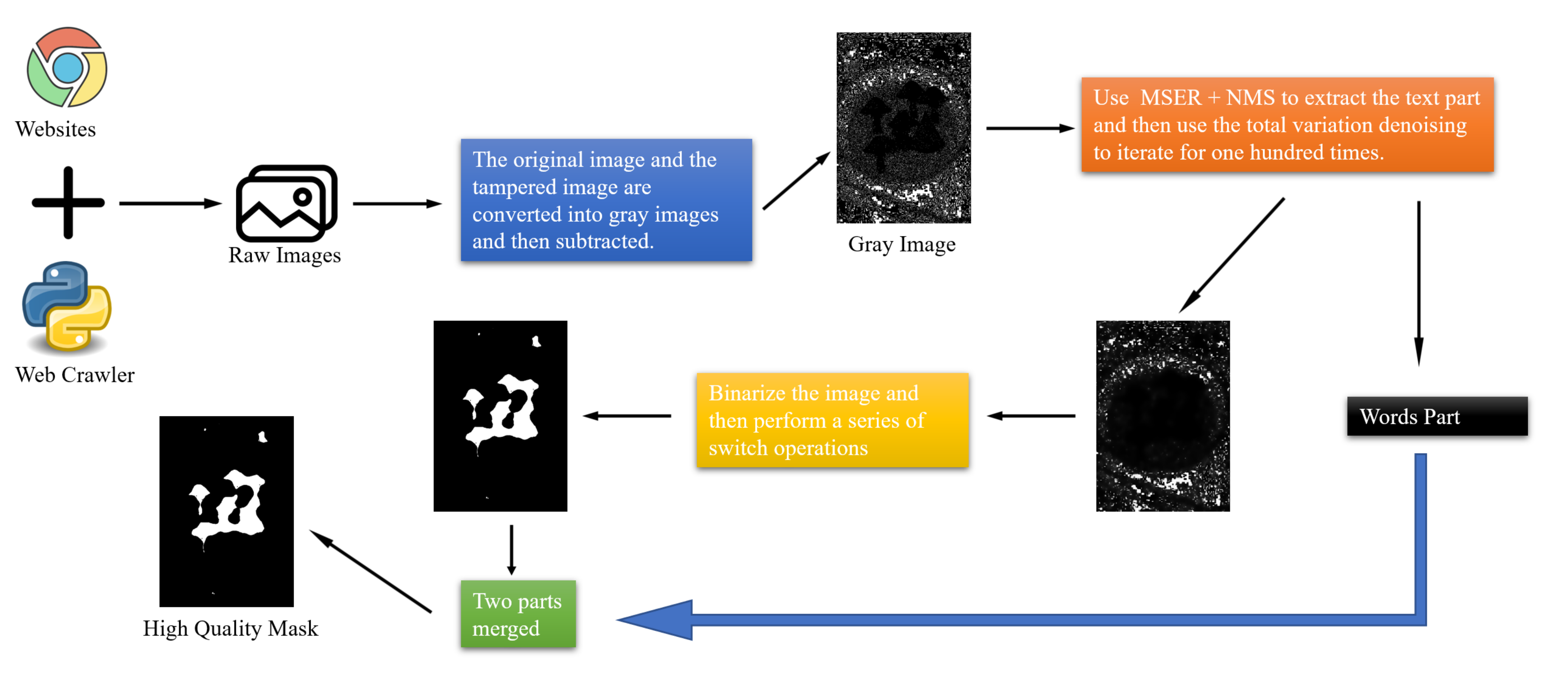}
\caption{ Modified total variation noise reduction architecture to obtain high-quality mask. Many original data is crawled from websites containing tampered images by crawlers. The original and tampered images are converted into gray images and then subtracted. Contour detection is used to extract the text part, and the gray image will be iterated one hundred times by total variation noise reduction. Then the image will be binarized and processed by switching operations. Words part and binary image add up to a high-quality mask.
}
\end{figure*}

\section{RELATED WORKS}

\subsection{Automated Data Crawling}
The dataset in the image tampered detection field is usually small because manual annotation is required. A growing wealth of information and increasingly sophisticated interfaces necessitate automated processing\cite{adc1} in recent years. The conventional web scraping methods for data collection include Web Crawlers, API Calls, RSS Subscriptions, and Database access. API, or Application Programming Interface, is a set of software tools that allows different applications to interact with each other. Through an API, we can obtain the desired data from other applications and return the data in a machine-readable format. RSS is a standard protocol for obtaining updates from blogs, news websites, and other sources. You can get updated content in real time by subscribing to RSS feeds. What we have applied is web crawlers. The web crawler is a program or software that traverses the Web and downloads web documents in a methodical, automated manner\cite{adc2}. Crawlers usually access the target website's pages according to predetermined rules, extract the data and save it locally or in a database. On the network, there are plenty of tampered images. We need to collect tampered images and their original images to get the data that could be used in the deep learning model. PS websites can meet this need. The PS bar in Baidu Post is our first choice.

\subsection{Total Variation}
The total variation has been introduced in Computer Vision by Rudin, Osher, and Fatemi\cite{tvb} as a regularizing criterion for solving inverse problems\cite{tv1}. It has many applications in image processing, computer vision, and image analysis. The total variation can be understood as the L1 norm of the gradient in the image, that is, the sum of the absolute values of the adjacent differences of the pixel values. The smaller the total variation, the higher the smoothness of the image. That is, the smaller the pixel value change in the image, and \textbf{\textit{vice versa}}. It means that there are more details and texture information in the image. In image processing, total variation is often used as a regularization term to constrain the solution of optimization problems, such as image noise reduction, image segmentation, image enhancement, etc. Total variation regularization can be achieved by adding a total variation term to the objective optimization function. Introducing the total variation regularization term can effectively suppress noise and protect image edge details from obtaining better image processing results. In addition to two-dimensional images, total variation can be extended to three-dimensional images, video, and other signal processing problems with wide application value. Arguably, the success of TV-based regularization lies in a good balance between the ability to model piece-wise smooth images and the difficulty of the resulting optimization problems\cite{tv2}.

\section{PROPOSED METHODS}

\subsection{Maximally Stable Extremal Regions} 
Our general working pipeline is depicted in Fig 2. For the commencement, we crawl tons of raw data from the website with tampered images. Only tampered images of the same size as the original can be processed. After the two images are subtracted and converted into a gray image, we will get an image with much noise. Usually, words in the image are obvious but thin. They are critical modified parts. However, processing them directly with total variation noise reduction technology will lead to the loss of this information. So we must pre-process the image with text recognition to retain this information. 

As the intensity contrast of text to its background is typically significant, and a uniform intensity or color within every letter can be assumed, MSER is a natural choice for text detection\cite{mser1}. The SIFT and SURF algorithms proposed by Lowe and Bay efficiently achieve feature detection with scale and rotation invariance, but these features are not affine-invariant. For various image regions with different shapes, affine-invariance is achieved by region rotation and size normalization. MSER is one of the most influential algorithms in region detection. The concept can be explained as follows. Imagine all possible three holdings of a gray-level image i. We will refer to the pixels below a threshold as 'black' and to those above or equal as 'white.' If we were shown a movie of threshold images, with the frame corresponding to a threshold, we would see first a white image. Subsequently, black spots corresponding to local intensity minimal will appear and grow. At some point, regions corresponding to two local minimal will merge. Finally, the last image will be black. The set of all connected components of all frames of the movie is the set of all maximal regions; minimal regions could be obtained by inverting the intensity of I and running the same process\cite{MSER}. Its mathematical principles are as follows: 

\begin{equation}
v(i)=\frac{\left|Q_{i+\Delta}-Q_{i-\Delta}\right|}{\left|Q_{i}\right|}\label{1}
\end{equation}

\subsection{Non-maximum Suppression}

We can put much-suspected text information in part of the box through MSER technology. However, dozens of boxes will be around the same part of the situation. These boxes overlap, which seriously interferes with our selection of the best effective region. So we use the Non-maximum Suppression method to choose the best one. Pedro F. Felzenszwalb\cite{nms} described an object detection system that represents highly variable objects using mixtures of multiscale deformable part models in 2009. These models are trained using a discriminative procedure that only requires bounding boxes for the objects in a set of images\cite{nms}. Their proposed technology is a component-based object detection system that uses a hybrid representation of multiscale deformable component models. In this detection system, NMS removes overlaps between multiple components. We take NMS to remove overlaps, too.

In target detection, NMS is a post-processing method for removing overlapped bounding boxes. After applying the bounding box prediction method described above, we have a set of detection D for a particular object category in an image. A bounding box and a score define each detection. We sort the detection in D by score and greedily select the highest-scoring ones while skipping detection with bounding boxes that are at least 50\% covered by a bounding box of a previously selected detection\cite{nms}. Through this theory, we can remove overlaps like Fig. 3

\begin{figure}[h]
\centering
\includegraphics[width=8cm]{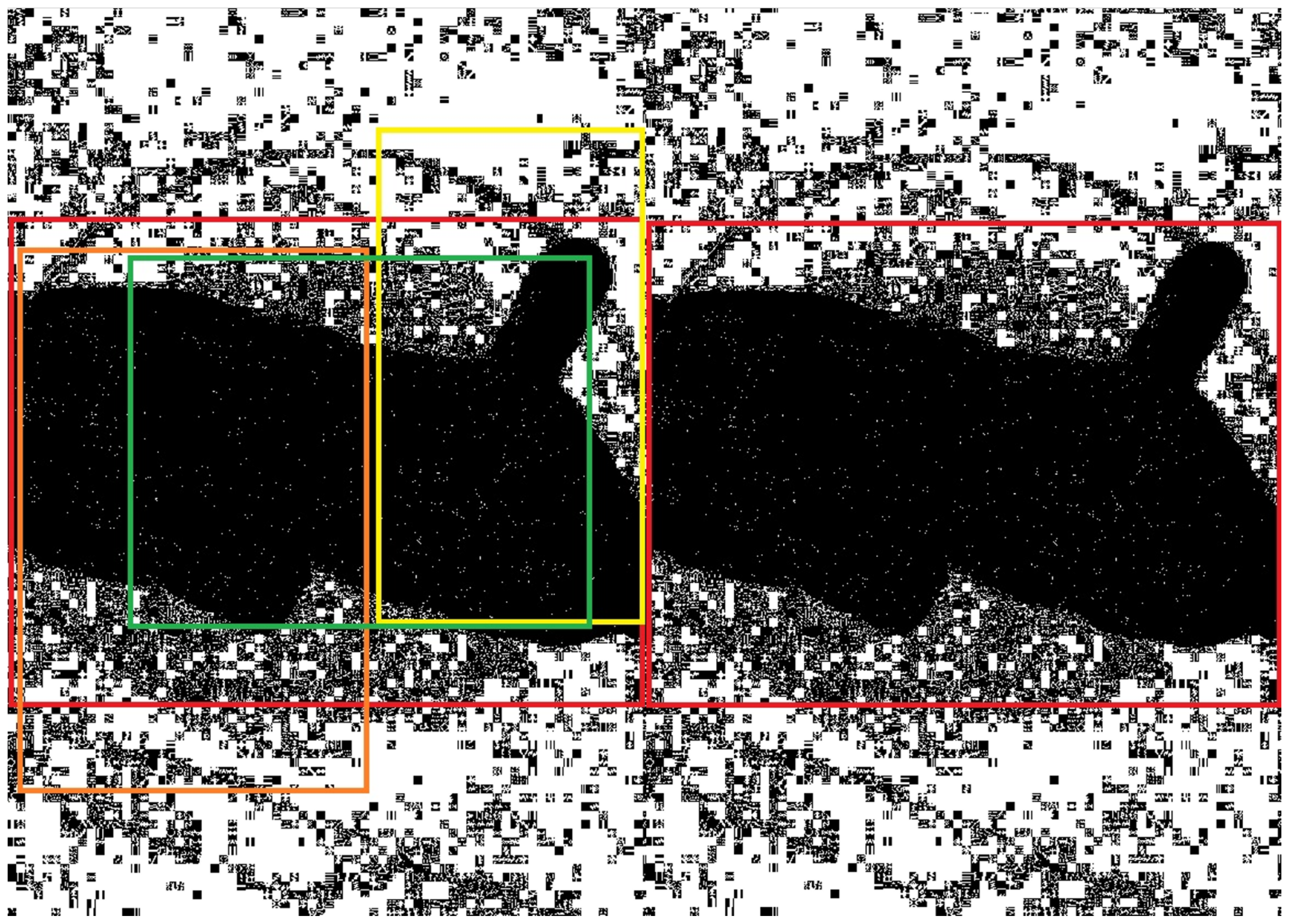}
\caption{The results of NMS. MSER will give us several rectangular frames, some of which are too large, some are too small, and some overlap too much, as shown in the left diagram. Our NMS technology reduces the multiple frames of the above into one of the most representative frames, as the right image shows.
}
\end{figure}

We can obtain many bounding boxes from MSER. In terms of evaluation metrics for bounding box regression, Intersection over Union (IoU) is the most popular metric\cite{IOU}.

\begin{equation}
IOU=\frac{A \cap B}{A \cup B}
\end{equation}

We set the intersection ratio $> = 0.4$ to delete the comparison block diagram, leaving the highest score block diagram. Go to the block diagram below the threshold, sort the remaining block diagram, select the block diagram with a high confidence value, and repeat the intersection and comparison process.

\subsection{Modified Total Variation Noise Reduction}
 Rudin et al.\cite{tvb} observed in 1990 that the total variation of noise-contaminated images is significantly more significant than that of noise-free images. Total variation (TV) methods are very effective for recovering “blocky,” possibly discontinuous, images from noisy data\cite{mtv1}. The total variation is defined as the integral of the gradient amplitude. Limiting the total variation limits the noise. The total variation of the image is minimized subject to constraints involving the noise statistics. The constraints are imposed using Lagrange multipliers. The solution is obtained using the gradient-projection method. This amounts to solving a time-dependent partial differential equation on a manifold determined by the constraints, as $t \rightarrow \infty$, the solution converges to a steady state which is the denoised image. The numerical algorithm is simple and relatively fast\cite{tvb}. In the data we obtain, the signal that may be false details has a high total variation. That is, the integral of the absolute gradient of the signal is high. According to this principle, the total change of the signal is reduced so that it is closely matched with the original signal. The unwanted details are removed while retaining important details, such as edges. The mathematical definition of the total variation is as follows:

 $$
 \inf _{u} I(u), \quad I(u):=\int_{\Omega}|\nabla u| \mathrm{d} x \mathrm{~d} y\label{2}
 $$

 However, in digital images, there is no continuous function. The whole image is composed of pixels, and they are discrete. We use pixel $x_{t}$ and the next pixel $x_{t+1}$ to represent $x_{t+1}\rightarrow x_{t}$. The formula is modified like this:

 $$
 V(y)=\sum_{i, j} \sqrt{\left|y_{i+1, j}-y_{i, j}\right|^{2}+\left|y_{i, j+1}-y_{i, j}\right|^{2}} \label{3}
 $$

In this way, we can process the masks which have massive noise. For the processed image, we calculate each pixel's first-order and second-order partial derivatives in the x and y directions by looping. To make the difference between the result and the original image not too large, we add a fidelity term when solving this gradient minimum. The value of each point is calculated according to the formula, and then we iterate the above operation one hundred times. The final image still can't be directly used because it's even more blurred. Using two erosion operations on this image, then binarizing it with a threshold of 15, we can get an image that has less noise. There are still some sporadic pixels we don't need. We can get an image with far less noise by processing them with 8 erosion operations and 2 dilation operations. Besides, inputting different values to the constants in the fidelity term will result in different processed images. Adding up the white parts of the total variation denoising image and the same image processed by MSER, we can obtain the high quality which could be used in the deep learning model.

\section{EXPERIMENTS AND RESULTS}
\subsection{Maximally Stable Extremal Regions Experiments and Results}

Experiments are conducted on the tampered image dataset crawled from Baidu Post. The dataset has more than 45000 images. We've grouped images that crawled down from the same post into one category. The first image is regarded as an original image by us. The left images are almost tampered images based on it. We throw out images that aren't the same size as the original, for most of these images are irrelevant emoticons. MSER can help us easily box out the words in the mask. There are 71 bounding boxes before NMS. And there are 41 bounding boxes after NMS. We can see the result in Fig.4

\begin{figure}[h]
\centering
\includegraphics[width=6cm]{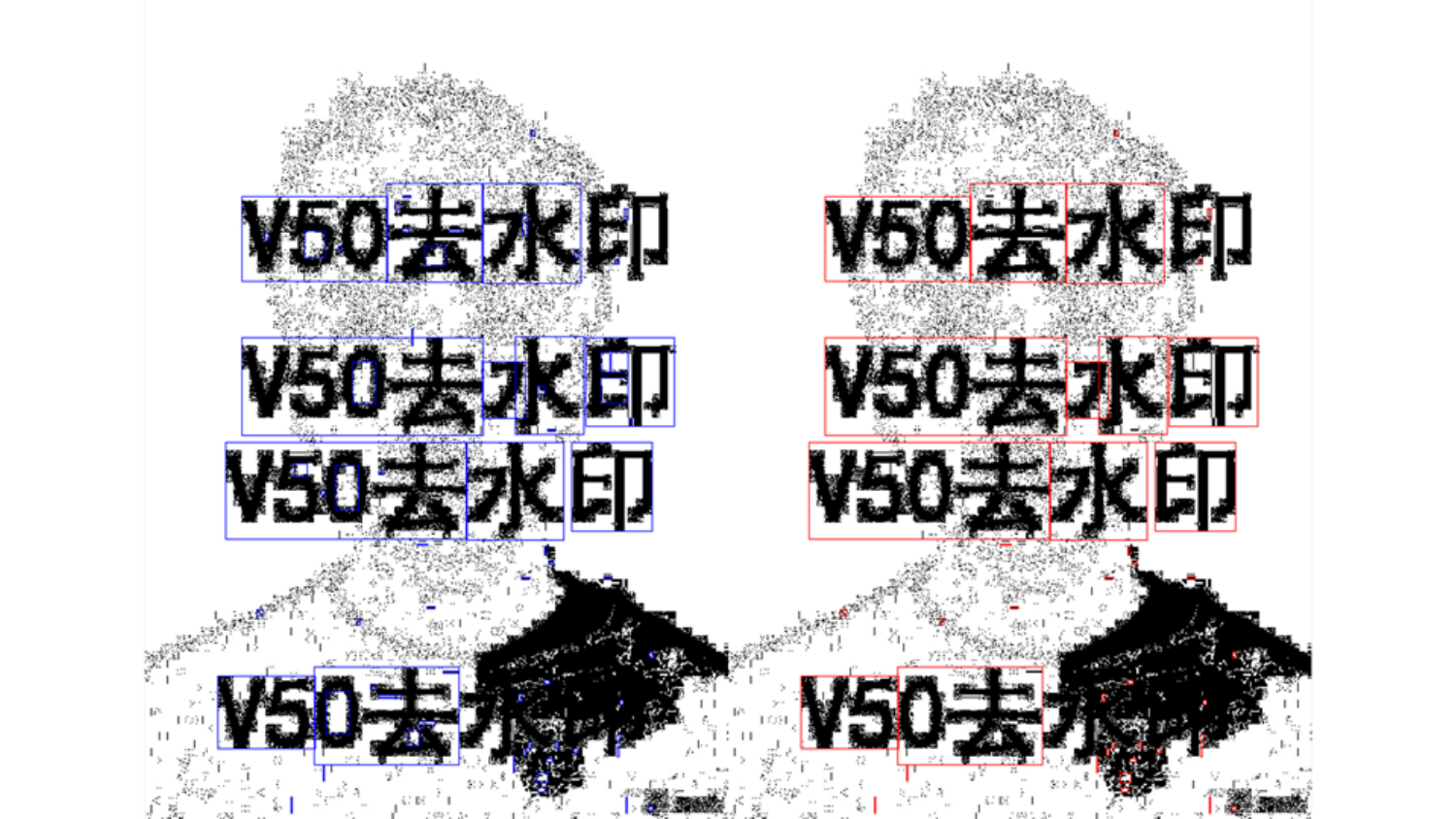}
\caption{MSER can accurately frame the part with the same gray value, most of which are text, and a few parts that are not text are other contents that have been tampered with. We can see 71 bounding boxes in the left graph. After NMS, there are only 41 boxes, as the right shows. The meaning of Chinese characters: pay 50¥ to remove the watermark.
}
\end{figure}

The words we boxed out are visible. In OCR image text recognition, most of the characters can be identified. Because of the lack of tampered images with words, we randomly selected 50 images to detect OCR text and make the results into a table. The text size in the picture is divided into five categories in Fig. 5. 

\begin{figure}[h]
\centering
\includegraphics[width=6cm]{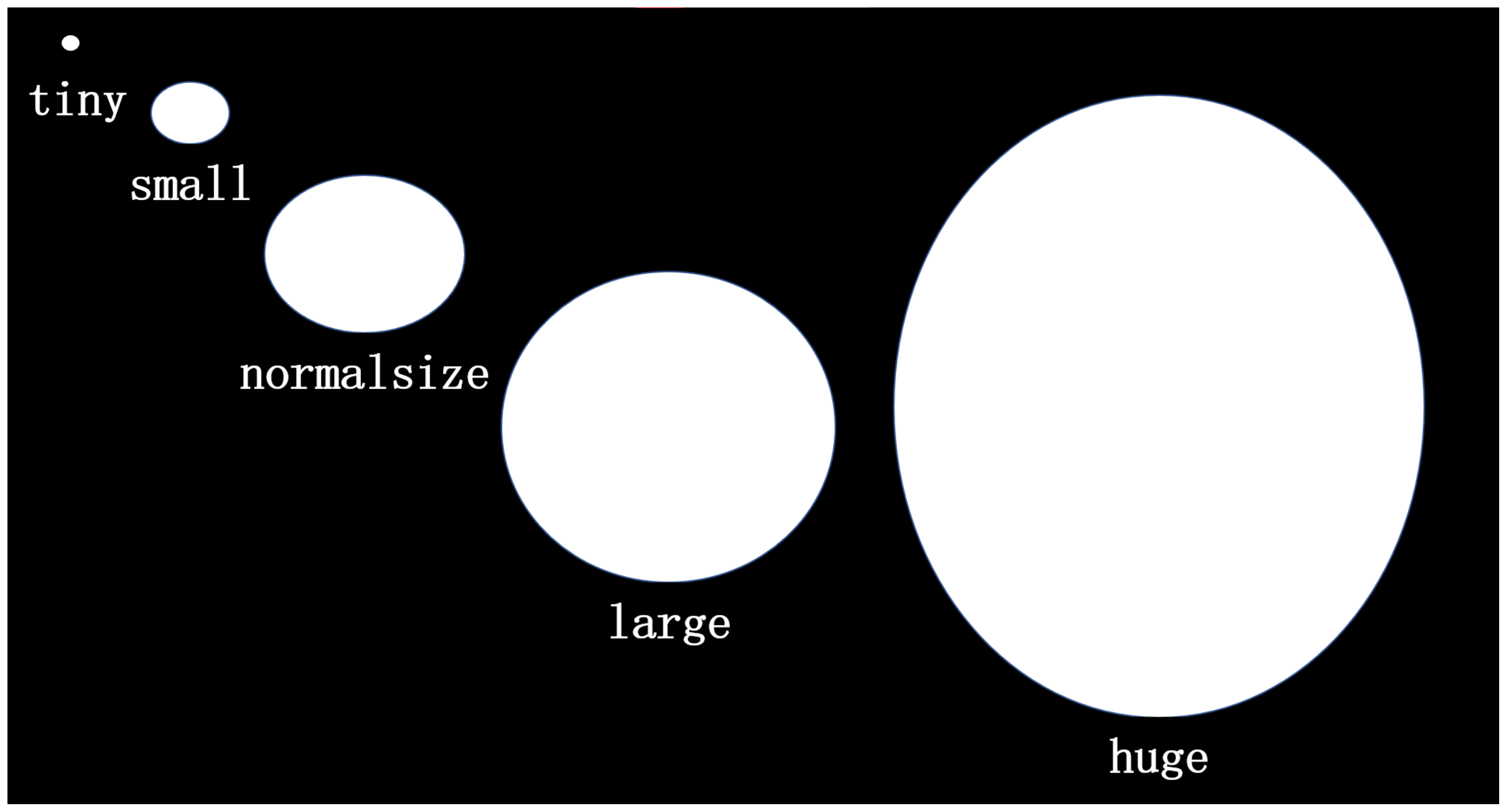}
\caption{The size of the character. Different font sizes will be affected to varying degrees in the subtraction of the two images. As shown above, we divide the font size into five specifications and calculate the accuracy of our method in different font sizes.
}
\end{figure}

We will show the rate of text recognition in the following table II. 

\begin{table}[!htbp]
\normalsize
    \caption{Text recognition rate}\label{tab:001} \centering
    \begin{tabular}{ccccc}
    
        \toprule[1.5pt]
        Body Size                  & Recognition Rate\\
        \midrule[1pt]
        tiny                 & 40.3\%   \\
        small                  & 64.7\%  \\
        normalsize                   & 85.4\%   \\
        large                    & 91.2\%  \\
         huge                      & 92.4\%  \\

        \bottomrule[1.5pt]
    \end{tabular}
\end{table}

The tiny and small section rate is low because, in the tampered graph, small words are greatly affected by tampering. This kind of word will be significantly deformed after the subtraction operation, resulting in not being recognized by OCR. For larger fonts, only a few extremely deformed characters can not be recognized. But all the characters are converted into white parts in the final mask. Even the fonts that cannot be identified can be marked by our algorithm.

\subsection{Modified Total Variation Noise Reduction Experiments and Results}

After extracting the required text information, we continue to perform total variation noise reduction on the original image. We combine the graphic part obtained by the total variation noise reduction processing with the text part obtained by the MSER processing. The results were corroded twice to obtain the final results. We deploy the program on our server for 24-hour automatic image acquisition and processing to get high-quality masks. Part of the final results are shown in Figure 6.

\begin{figure}[h]
\centering
\includegraphics[width=3.5cm,height=5cm]{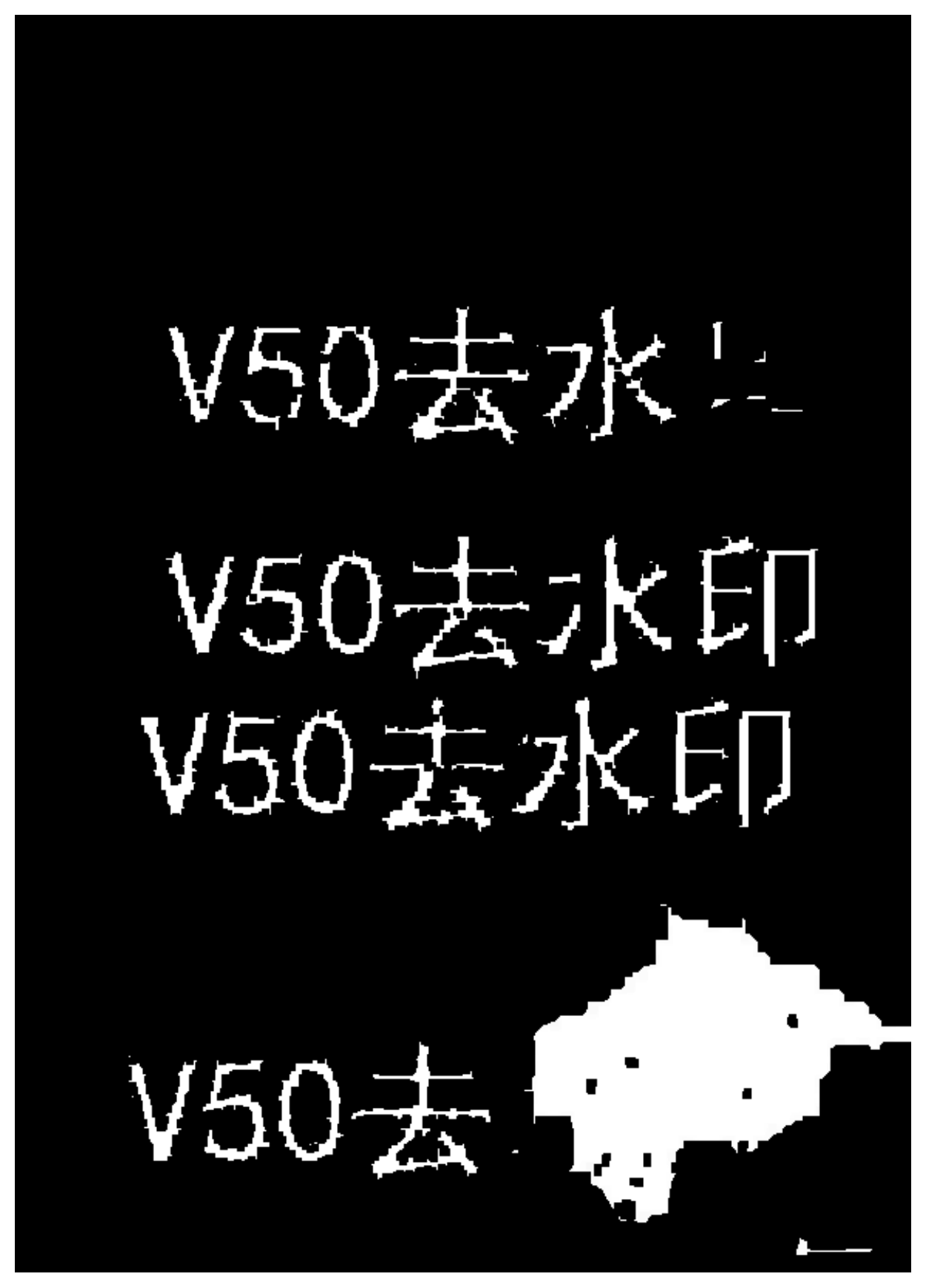}
\caption{The final result graph was obtained by combining the text part and the graphic part. The meaning of Chinese characters: pay 50¥ to remove the watermark.
}
\end{figure}

Obviously, our operation greatly reduces the noise and retains the text information so that the deep learning model can use the mask.

\subsection{Advantages compared to existing methods}

\begin{itemize}
\item Large-Scale. Our raw data contains 46509 images, while the commonly used tampering detection dataset CASIA v1.0, CASIA v2.0, Columbia, COLUMB, FORENSICS only contains 1725, 12323, 1845, 358, and 288 images, respectively. Such a larger scale can help better exploit the full potentials of more advanced model architectures\cite{data_good}.
\item Continuous Regeneration. Commonly used data uses random Gaussian noise, JPRG compression, and random flipping to make the data set one more, restricting the increase in the data. On the contrary, net friends could supply countless raw data for us. What we need to do is crawl their images on the internet.
\item Covering a Wide Range. Many datasets have their particular limitation: The splicing forgery region of the CASIA data set is a small and delicate object. The splicing forgery region of COLUMB is superficial, large, and meaningless. But the dataset we build has a wide range. The types of tamper depend on how many netizens can offer.
\item High-Speed Generation. Our experiments are on a 16g RAM, i7-11800H CPU, RTX 3060 GPU machine. An average size image takes around 3 minutes to process into our mask. The large image needs more time, and the small image only needs less than 1 minute. In this way, we can automatically get a lot of data in a very short time. Our work efficiency will increase exponentially if multiple devices are processed in parallel.

\end{itemize}

\section{CONCLUSIONS}

In this paper, we build the Modified Total Variation Noise Reduction architecture to automatically crawl raw data from the Internet and make it a high-quality mask for deep learning. This technology that can obtain many high-quality masks quickly is expected to help our deep learning model achieve better learning results. We want to apply this technology to tampering detection, which requires many datasets. If the deep learning model has enough data, the harm of tampering with images in the future may be significantly reduced.

\section{ACKNOWLEDGEMENT}

Numerical computations of this paper are supported by Hefei advanced computing center.














\begin{thebibliography}{10}
\providecommand{\url}[1]{#1}
\csname url@samestyle\endcsname
\providecommand{\newblock}{\relax}
\providecommand{\bibinfo}[2]{#2}
\providecommand{\BIBentrySTDinterwordspacing}{\spaceskip=0pt\relax}
\providecommand{\BIBentryALTinterwordstretchfactor}{4}
\providecommand{\BIBentryALTinterwordspacing}{\spaceskip=\fontdimen2\font plus
\BIBentryALTinterwordstretchfactor\fontdimen3\font minus \fontdimen4\font\relax}
\providecommand{\BIBforeignlanguage}[2]{{%
\expandafter\ifx\csname l@#1\endcsname\relax
\typeout{** WARNING: IEEEtran.bst: No hyphenation pattern has been}%
\typeout{** loaded for the language `#1'. Using the pattern for}%
\typeout{** the default language instead.}%
\else
\language=\csname l@#1\endcsname
\fi
#2}}
\providecommand{\BIBdecl}{\relax}
\BIBdecl

\bibitem{nms}
P.~F. Felzenszwalb, R.~B. Girshick, D.~McAllester, and D.~Ramanan, ``Object detection with discriminatively trained part-based models,'' \emph{IEEE Transactions on Pattern Analysis and Machine Intelligence}, vol.~32, no.~9, pp. 1627--1645, 2010.

\bibitem{MSER}
J.~Matas, O.~Chum, M.~Urban, and T.~Pajdla, ``\BIBforeignlanguage{en}{Robust {Wide} {Baseline} {Stereo} from {Maximally} {Stable} {Extremal} {Regions}}.''

\bibitem{tvb}
\BIBentryALTinterwordspacing
L.~I. Rudin, S.~Osher, and E.~Fatemi, ``\BIBforeignlanguage{en}{Nonlinear total variation based noise removal algorithms},'' \emph{\BIBforeignlanguage{en}{Physica D: Nonlinear Phenomena}}, vol.~60, no. 1-4, pp. 259--268, Nov. 1992. [Online]. Available: \url{https://linkinghub.elsevier.com/retrieve/pii/016727899290242F}
\BIBentrySTDinterwordspacing

\bibitem{background1}
P.~Zhou, X.~Han, V.~I. Morariu, and L.~S. Davis, ``Learning rich features for image manipulation detection,'' in \emph{Proceedings of the IEEE Conference on Computer Vision and Pattern Recognition (CVPR)}, June 2018.

\bibitem{background2}
C.~Beijing, J.~Xingwang, G.~Ye, and W.~Jinwei, ``A quaternion two-stream r-cnn network for pixel-level color image splicing localization,'' \emph{Chinese Journal of Electronics}, vol.~30, no.~6, pp. 1069--1079, 2021.

\bibitem{background3}
S.~Zhang, H.~Tong, J.~Xu, and R.~Maciejewski, ``Graph convolutional networks: a comprehensive review,'' \emph{Computational Social Networks}, vol.~6, no.~1, pp. 1--23, 2019.

\bibitem{adc1}
T.~Furche, G.~Gottlob, G.~Grasso, C.~Schallhart, and A.~Sellers, ``Oxpath: A language for scalable data extraction, automation, and crawling on the deep web,'' \emph{The VLDB Journal}, vol.~22, pp. 47--72, 2013.

\bibitem{adc2}
M.~A. Kausar, V.~Dhaka, and S.~K. Singh, ``Web crawler: a review,'' \emph{International Journal of Computer Applications}, vol.~63, no.~2, pp. 31--36, 2013.

\bibitem{tv1}
A.~Chambolle, ``An algorithm for total variation minimization and applications,'' \emph{Journal of Mathematical imaging and vision}, vol.~20, pp. 89--97, 2004.

\bibitem{tv2}
J.~P. Oliveira, J.~M. Bioucas-Dias, and M.~A. Figueiredo, ``Adaptive total variation image deblurring: a majorization--minimization approach,'' \emph{Signal processing}, vol.~89, no.~9, pp. 1683--1693, 2009.

\bibitem{mser1}
H.~Chen, S.~S. Tsai, G.~Schroth, D.~M. Chen, R.~Grzeszczuk, and B.~Girod, ``Robust text detection in natural images with edge-enhanced maximally stable extremal regions,'' in \emph{2011 18th IEEE international conference on image processing}.\hskip 1em plus 0.5em minus 0.4em\relax IEEE, 2011, pp. 2609--2612.

\bibitem{data1}
J.~Dong, W.~Wang, and T.~Tan, ``Casia image tampering detection evaluation database,'' in \emph{2013 IEEE China summit and international conference on signal and information processing}.\hskip 1em plus 0.5em minus 0.4em\relax IEEE, 2013, pp. 422--426.

\bibitem{mtv1}
``Iterative methods for total variation denoising,'' \emph{SIAM Journal on Scientific Computing}, vol.~17, no.~1, pp. 227--238, 1996.

\bibitem{background5}
Y.~Zhang, J.~Goh, L.~L. Win, and V.~L. Thing, ``Image region forgery detection: A deep learning approach.'' \emph{SG-CRC}, vol. 2016, pp. 1--11, 2016.

\bibitem{background6}
D.~Tralic, I.~Zupancic, S.~Grgic, and M.~Grgic, ``Comofod—new database for copy-move forgery detection,'' in \emph{Proceedings ELMAR-2013}.\hskip 1em plus 0.5em minus 0.4em\relax IEEE, 2013, pp. 49--54.

\bibitem{IOU}
Z.~Zheng, P.~Wang, W.~Liu, J.~Li, R.~Ye, and D.~Ren, ``Distance-iou loss: Faster and better learning for bounding box regression,'' in \emph{Proceedings of the AAAI conference on artificial intelligence}, vol.~34, no.~07, 2020, pp. 12\,993--13\,000.

\bibitem{dataset0}
G.~Mahfoudi, B.~Tajini, F.~Retraint, F.~Morain-Nicolier, and M.~Pic, ``Defacto: Image and face manipulation dataset,'' in \emph{2019 27th European Signal Processing Conference (EUSIPCO)}, 2019.

\bibitem{datasets}
\BIBentryALTinterwordspacing
Q.~Yang, D.~Chen, Z.~Tan, Q.~Liu, Q.~Chu, J.~Bao, L.~Yuan, G.~Hua, and N.~Yu, ``\BIBforeignlanguage{en}{{HQ}-{50K}: {A} {Large}-scale, {High}-quality {Dataset} for {Image} {Restoration}},'' Jun. 2023, arXiv:2306.05390 [cs]. [Online]. Available: \url{http://arxiv.org/abs/2306.05390}
\BIBentrySTDinterwordspacing

\bibitem{data_good}
\BIBentryALTinterwordspacing
------, ``\BIBforeignlanguage{en}{{HQ}-{50K}: {A} {Large}-scale, {High}-quality {Dataset} for {Image} {Restoration}},'' Jun. 2023, arXiv:2306.05390 [cs]. [Online]. Available: \url{http://arxiv.org/abs/2306.05390}
\BIBentrySTDinterwordspacing

\end{thebibliography}

\end{document}